\documentclass{article}

\usepackage{amsmath,amsfonts,amssymb,amscd,amsthm,xspace}
\usepackage{xcolor}
\usepackage{graphicx}
\usepackage{subcaption}
\usepackage{hyperref}
\usepackage{algorithm}
\usepackage{algpseudocode}

\theoremstyle{definition}

\theoremstyle{definition}
\newtheorem{remark}{Remark}[section]

\theoremstyle{definition}

\DeclareMathOperator*{\argmax}{arg\,max}

\title{Understanding and contextualising  diffusion models}
\author{Stefano Scotta and Alberto Messina\\ \textit{RAI - Radiotelevisione Italiana}\\ \textit{Centro Ricerche, Innovazione Tecnologica e Sperimentazione}}
\begin{document}
\maketitle

\begin{abstract}
The latest developments in Artificial Intelligence include diffusion generative models, quite popular tools which can produce original images both unconditionally and, in some cases, conditioned by some inputs provided by the user. Apart from implementation details, which are outside the scope of this work, all of the main models used to generate images are substantially based on a common  theory which restores a new image from a completely degraded one. In this work we explain how this is possible by focusing on the mathematical theory behind them, i.e. without analyzing in detail the specific implementations and related methods.  The aim of this work is to clarify to the interested reader what all this means mathematically and intuitively.
\end{abstract}

\section{Introduction}
The main goal of this work is to give an as much rigorous as possible explanation of the theory behind the main models used nowadays for image generation. We talk about ``generative'' models because they are able, once opportunely set and trained, to generate new and realistic visual contents (meaning that they could look ``real'' images to an average observer). The common idea behind these models is: since we are able to corrupt gradually any real image up to a completely noisy one, then we should be able to do the reverse process of ``denoising'' a totally random noisy image. Hence, this process of generating new images can be intended as the reverse of the process that is used to corrupt the real images. In some way, the models that we are going to present, given a lot of sequences of images that are gradually corrupted, become able to recover, from a randomly sampled input corresponding to a completely corrupted image, a new realistic one (which is none of the original images used for training). The reference these models do to the concept of ``diffusion'' comes from the process used to corrupt the images as well as - we will see - from the one used to recover them. Indeed, in most of the literature about generative models the real images used for training are corrupted adding gradually, and for a really large number of steps, infinitesimal Gaussian noise to their initial representation. ``Noising'' processes built in this way are examples of diffusion processes. 

\begin{figure}
\begin{subfigure}{\textwidth}
  \centering
  \includegraphics[width=\linewidth, trim={1cm 1cm 0cm 0cm}]{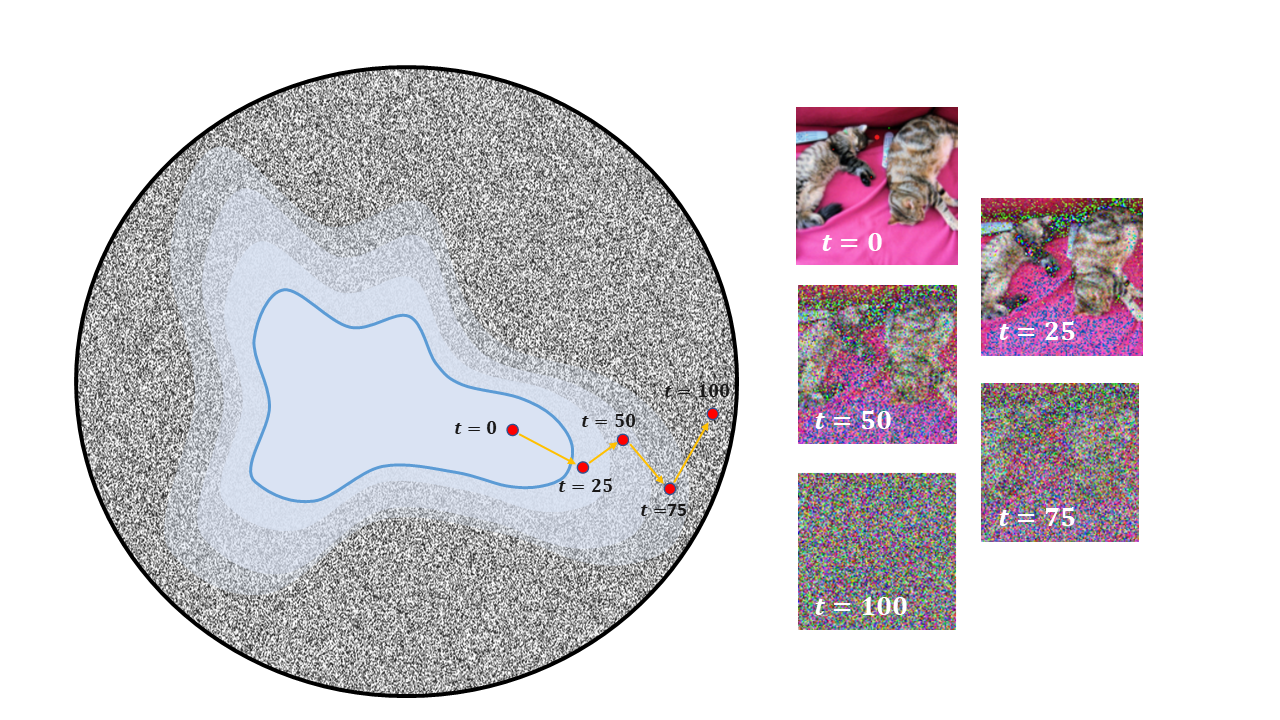}
  \caption{Forward ``noising'' process.}
  \label{fig:sfig1}
\end{subfigure}
\begin{subfigure}{\textwidth}
  \centering
  \includegraphics[width=\linewidth, trim={1cm 1cm 0cm 0cm}]{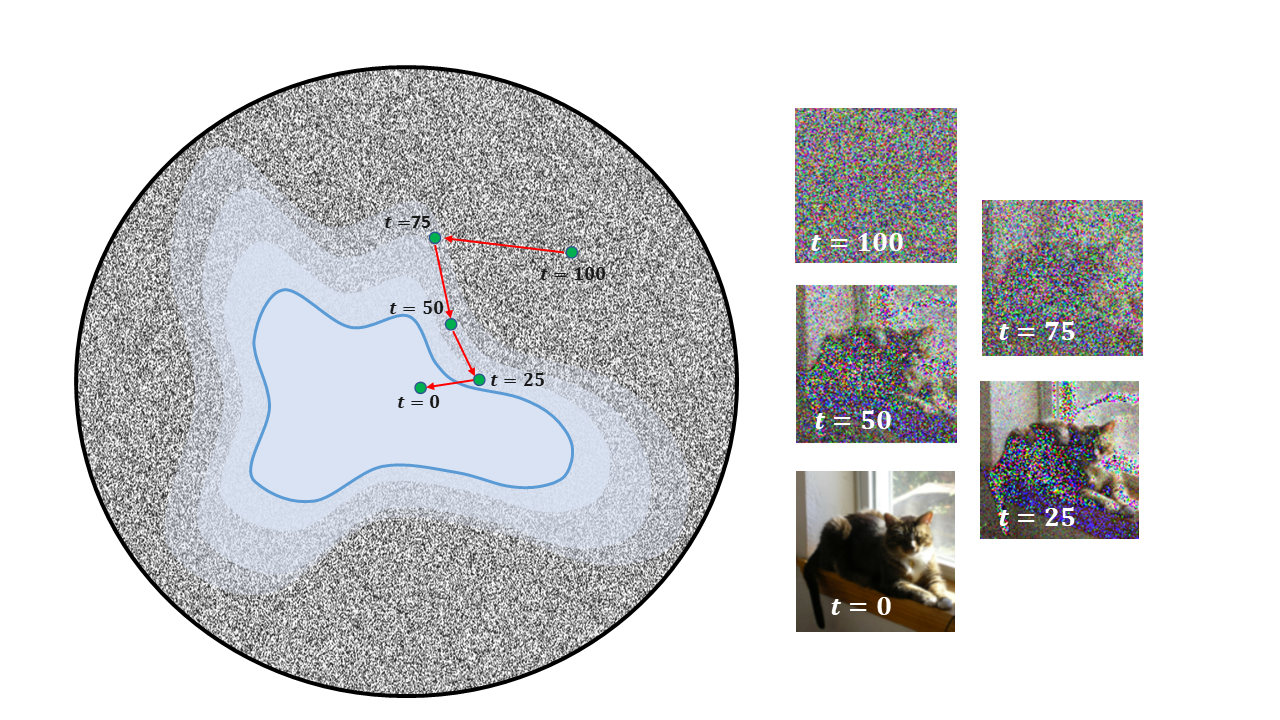}
  \caption{Backward ``denoising'' process.}
  \label{fig:int}
\end{subfigure}
\caption{Intuitive representation of the process adding noise starting from a given image (a) and the one denoising a completely noisy image to get to a new one. In blue we represented the set composed by all the real images starting point of the noising process and the limit point of the denosing process. Notice that this is just an intuitive representation of these processes, since they actually take place in much higher dimensions and we do not know anything about the properties of the spaces involved.}
\label{fig:fig}
\end{figure}

More in detail, any image is represented by a  precise sequence of numbers corresponding to the pixel values from which it can be rendered, and this sequence is corrupted, adding Gaussian noise, until it becomes a totally random sequence belonging to some probability distribution. This process is repeated many times and used to train the model in such a way that it will be also usable to compute the reverse process. The idea is then to take a random sample from the resulting probability distribution and use this model to obtain a new sequence of numbers corresponding to a new ``real'' image. These processes are intuitively represented in Figure \ref{fig:fig}.

In other words these models are able to sample new elements from the probability distribution of the sequences of pixel values representing real images, without explicit knowledge of this distribution. Elements of this distribution are the limit points of the reverse process described above, starting from any completely noisy image belonging to the limit distribution of the ``noising'' process. 

This approach is completely different from the one   trying to describe that distribution as a mixture of known probability distributions (for example Gaussian Misture Models) in order to be able to sample from that later. Indeed, even if this last approach is perfectly working in theory  (\cite{book}), it came out that the goal distribution is too complex to be obtained as a combination of a reasonable number of known probability distributions. A good reference for these attempts is the paper \cite{GMM} where some solutions are also proposed.

As anticipated, we decided to focus rather on the mathematical theory behind all the models that we are going to present than on the way in which they are being practically developed. Our goal is indeed to make understandable more the ``why'' this kind of model work than the ``how''.
Nevertheless, we hope that after reading this paper the interested reader can have a fresh look at the original papers and recognize in each of the proposed structures the theoretical foundation here described. 

This work is structured as follows. In Section \ref{diffusion_paper} we considered  the Diffusion Probabilistic Model (DPM), for which we focused particularly on the references \cite{termo, denoising}, the seminal works where the possibility to get a totally new image from a noisy image through modeling the reverse Guassian process (``denoising'' ) was firstly introduced. This model and the theory on which it is based are the key also to understand the models that later improved its performance, as the ones presented in \cite{denoising2, improved, bilateral, nongaussian} and many others. 
Then, in Section \ref{stable_sec} we briefly state the differences between DPM and the Stable Diffusion (DS) models introduced in \cite{stable}. This new (state of the art) kind of models improves the results and decreases the computational cost of DPM by using the same idea except for changing the space in which it works, namely a space of latent representations of images. Moreover, it also allows to condition the generation process through the insertion of a prompt influencing the kind of image we want to generate.
Lastly, in Section \ref{cold_sec}, we talk about the recent proposals for Cold Diffusion models (developed in \cite{cold}), that use the same idea behind DPM but proving that it is not actually necessary to use random elements to corrupt and then restore the images. Indeed the authors showed that using arbitrary corrupting processes (including  deterministic ones) it is possible to build a model that can generate new images starting from a totally corrupted one. 

\subsection{Similar works} 
Because of the novelty introduced by these kind of processes during the last months, many works similar to this one have been published with the aim to explain the theory behind generative models and in particular behind DPM. \cite{google} surely deserves a particular mention, since indeed in that work the whole mathematical basics behind DPM are treated in details similarly as we did here. Other surveys (like \cite{survey, survey2}) and many web blogs also tried to make these kind of models more understandable to the occasional reader. We decided anyway to complete this work because we believe that it contains some small novelties and some additional comments that can be useful to anyone interested in understanding better the underlying theory.

\section{Diffusion Probabilistic Models}\label{diffusion_paper}
In this section we present the DPM introduced in \cite{denoising, termo}. The main idea behind this model, as we mentioned in the Introduction, is that, given the stochastic process adding noise to real images up to a completely noisy image, it is possible to reverse this process, i.e., starting from a completely noisy image, to denoise it, step by step, up to an image belonging to the  distribution of realistic images and - as such - that looks totally realistic. This idea implies that the built model, in some way, ``compiles'' the stochastic relation between the parts (pixels) of the noisy image and uses this compiled knowledge to recover (in a stochastic sense) some combination of pixels representing a realistic image. Furthermore, we can say that the resulting image is (with some abuse of notation, see Remark \ref{prob1}) a totally new image with probability 1. Intuitively, this conjecture is based on the observation that because the space of all possible realistic images is (almost) infinite - and since the reverse process starting from a random pattern goes back to an image belonging to the distribution of real images in a random way -, the limiting point is ``almost surely'' different from each one that actually already ``exists'' (obviously including the image set used in the training process). 

\begin{remark}\label{prob1}
We said that the images generated by these models are ``new'' with probability 1 but, even if this is true in practice, it is not true theoretically. In fact, since the images that we consider are composed by a finite number of pixel values combinations, they are not infinite. Therefore, theoretically, any of the real images (even what you are seeing right now) could be reproduced by these models, the limit being just in the amount and diversity of the training data. However, in practice the set of known realistic images that each of us observes (let alone remembers) during its whole life is so limited compared to the set of all possible realistic images that this never happens. 
\end{remark}

\subsection{Mathematical background}\label{background}
The idea of the forward process involved here is to add Gaussian noise to a real image $x_0$ at each step up to a completely noised image. In particular this process is a Markov process $\{x_t\}_{t \in \{0,\dots,T\}}$ (defined on some probability space $(\Omega, \mathcal{F},\mathbb{P})$) taking values in the set of all possible images, meant as matrices containing the values composing the image\footnote{Hitherto and for the remainder of the paper we consider images, also in order to give the reader a concrete reference with which to follow the tractation, but notice that the whole theory does not depend on the space in which $\{x_t\}_{t \in \{0,\dots,T\}}$ is in, as it comes out from Section \ref{stable_sec}.}. Given the starting image $x_0$, the forward ``noising'' process is built iterating the following step in which we add Gaussian noise for any $t \in \{1,\dots, T\}$:
\begin{equation}\label{process}
x_t=\sqrt{1-\beta_t}x_{t-1}+\sqrt{\beta_t}z_{t},
\end{equation}
where $z_{t} \sim \mathcal{N}(0,I)$ and $\{\beta_t\}_{t=1}^T$ are arbitrary positive parameters.
So, the distribution of $x_t$ given $x_{t-1}$, that we denote by $p(x_t|x_{t-1})$ is the same of a random variable $\mathcal{N}(\sqrt{1-\beta_t}x_{t-1}; \beta_t I)$, for any $t \in \{1,\dots, T\}$.
%(we will use the notation $p(x_{t}|x_{t-1})=\mathcal{N}(\sqrt{1-\beta_t}x_{t-1}; \beta_t I)$ hereinafter).

Moreover, being the process Markovian, with these conditional distributions we can explicitly evaluate the distribution of the whole process, i.e.
\begin{equation}\label{markov1}
p(x_0,x_1, \dots, x_T)=p(x_0)p(x_1, \dots, x_T|x_0)=p(x_0)\prod_{t=1}^T p(x_t|x_{t-1}).
\end{equation} 

Another interesting and useful property of this kind of process is that, given $x_0$ it is possible to evaluate the distribution of $x_t$, for any $t \in \{1,\dots, T\}$, as follows
\begin{equation}
\begin{split}
x_t&=\sqrt{1-\beta_t}x_{t-1}+\sqrt{\beta_t}z_t=\sqrt{1-\beta_t}(\sqrt{1-\beta_{t-1}}x_{t-2}+\sqrt{\beta_{t-1}}z_{t-1})+\sqrt{\beta_t}z_t\\
&=x_0\prod_{s=1}^{t}\sqrt{1-\beta_t}+\sum_{s=1}^{t}\sqrt{\beta_s}z_s\prod_{u=s+1}^t \sqrt{1-\beta_{u}}.
\end{split}
\end{equation}
Hence, introducing the notation $\alpha_t=1-\beta_t$ and $\overline \alpha_t=\prod_{s=1}^t \alpha_s$, we have $p(x_t|x_0)=\mathcal{N}(x_0\sqrt{\overline \alpha_t};(1-\overline \alpha_t)I)$ (see Remark \ref{variance} for details about it). Then let us observe two very important features that this process has if we opportunely choose the values of the $\{\beta_t\}_{t=1}^T$. 
\begin{itemize}
\item if $\overline \alpha_t \rightarrow 0$ as $t \rightarrow \infty$, the limit distribution of $x_t$ is a $\mathcal{N}(0;I)$. This means that, under this hypothesis, if we take sufficiently big $T$ , we can say that $x_T \sim \mathcal{N}(0;I)$, independently from the starting point $x_0$. Note that $\overline \alpha_t \rightarrow 0$ as $t \rightarrow \infty$, if, for example, $\beta_s<1$ for any $s$. This property is clear from the simulation results resumed in Figure \ref{fig:example}.

\item if $\beta_t <<1$ for any $t \in \{1,\dots,T\}$, as pointed out in \cite{termo} and in Remark \ref{diffusion}, the process $\{x_t\}_{t=0}^T$ admits a reverse process $\{x_t\}_{t=T}^0$ such that the conditional distributions $p(x_{t-1}|x_t)$ have the same functional form of the forward ones $p(x_t|x_{t-1})$, for any $t \in \{1,\dots, T\}$. This is really the key feature of this process in order to characterize the reverse process in the following sections.

\end{itemize} 

\begin{remark}
Notice that if we take all values of $\{\beta_{t}\}_{t=1}^T$ close to $1$ the process $x_t$ converges really fast to the limit distribution $\mathcal{N}(0,I)$. However, in this case we can not assume that there exists a reverse diffusion process with the same functional form (see Remark \ref{diffusion}) and so that the model presented would be able to recover the reverse process.
\end{remark}

\begin{figure}[ht]
  \includegraphics[width=\linewidth, trim={0cm 0cm 0cm 0cm}]{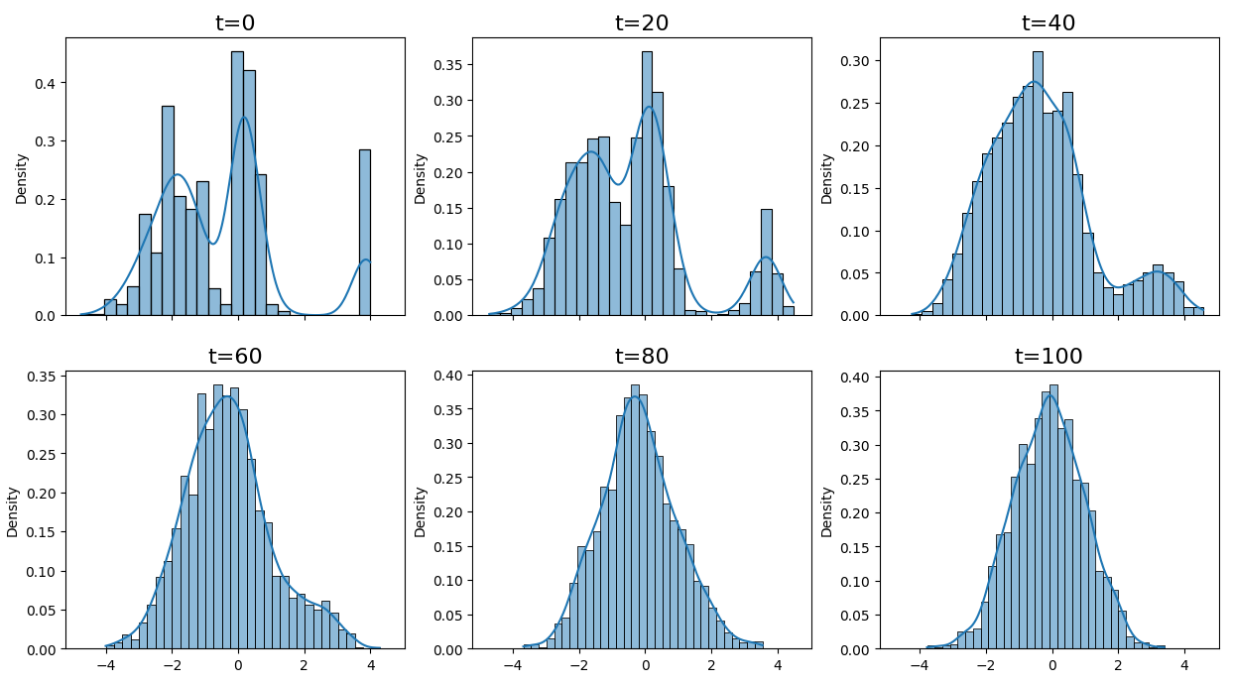}
  \caption{Distribution in time of the values of $x_t$ in one dimension, obtained starting from a given distribution $x_0$ (in the top left plot) after 2000 simulations of trajectories. The parameters $\beta_t$ were chosen growing linearly in time from $0.0004$ to $0.06$. It is clear the convergence to some distribution close to a Gaussian random variable after a sufficient big number of steps, as we proved analytically. Doing the same simulation with higher values of 
 $\{\beta_t\}_{t \in \{1,\dots,T\}}$ the convergence to the Gaussian distribution would be much faster (immediate in the case $\beta_t=1$ for any $t \in \{1,\dots,T\}$). The notebook containing the simulations related to this toy model and other interesting remarks can be found at \url{https://github.com/stefanoscotta/1-d-generative-diffusion-model}.}
  \label{fig:example}
\end{figure}

\begin{remark}\label{variance}
It can easily be proved by induction that the random variable $\sum_{s=1}^{t}\sqrt{\beta_s}z_s\prod_{u=s+1}^t\sqrt{1-\beta_{u}} \sim \mathcal{N}(0, (1-\prod_{s=1}^t(1-\beta_{s})I)= \mathcal{N}(0, (1-\overline \alpha_t)I)$. Indeed the sum of independent Gaussian random variable still a Gaussian r.v. with mean equal to the sum of the means and variance equal to the sum of the variances. Moreover, the variance of $\sqrt{\beta_s}z_s\prod_{u=s+1}^t\sqrt{1-\beta_{u}}$ is equal to $\beta_s\prod_{u=s+1}^t(1-\beta_{u})$ for any $s \in \{1,\dots, t\}$. Let us then show that the base case of the induction proof holds: for $t=2$, the variance of $\sum_{s=1}^{2}\sqrt{\beta_s}z_s\prod_{u=s+1}^2\sqrt{1-\beta_{u}}$ is equal to $\beta_1(1-\beta_2)+\beta_2=1-(1-\beta_1)(1-\beta_2)=1-\overline\alpha_2$. Therefore it remains only to show the inductive step to concludes the proof, let assume that for $t=t'$ it holds that the variance of  $\sum_{s=1}^{t'}\sqrt{\beta_s}z_s\prod_{u=s+1}^{t'}\sqrt{1-\beta_{u}}$ is $(1-\overline \alpha_{t'})$, then the variance of $\sum_{s=1}^{t'+1}\sqrt{\beta_s}z_s\prod_{u=s+1}^{t'+1}\sqrt{1-\beta_{u}}$ is equal to
\begin{equation*}
\begin{split}
\beta_{t'+1}+&\sum_{s=1}^{t'}\beta_s\prod_{u=s+1}^{t'}(1-\beta_u)(1-\beta_{t'+1})=\beta_{t'+1}+(1-\overline\alpha_{t'})(1-\beta_{t'+1})\\&=\beta_{t'+1}+1-\beta_{t'+1}-\overline\alpha_{t'}(1-\beta_{t'+1})=1-\overline\alpha_{t'+1}.
\end{split}
\end{equation*}
\end{remark}

\begin{remark}\label{diffusion}
We want to find the continuous version of \eqref{process} in order to use some important properties of this kind of processes. To do this we follow the computations made in Appendix B of \cite{score}. 
Under the condition $\beta_t <<1$, for any $t \in \{1,\dots, T\}$, we are able to show that we can approximate \eqref{process} as
\begin{equation}\label{diff1}
x_{t+1}=\sqrt{1-\beta_{t+1}}x_{t}+\sqrt{\beta_{t+1}}z_{t+1} \approx \Big(1-\frac{\beta_{t+1}}{2}\Big)x_{t}+\sqrt{\beta_{t+1}}z_{t+1}.
\end{equation}
Indeed, developing the Taylor series of $\sqrt{1-\beta_{t+1}}$ around $0$ we get that it is equal to $1-\frac{\beta_{t+1}}{2}$ plus a term of order $\beta_{t+1}^2$ that can be ignored.

Let us now introduce a new set of scaling parameters $\{\tilde \beta_t=T\beta_t\}_{t=1}^T$ with which we can rewrite \eqref{diff1} as 
$$
x_{t+1}-x_t\approx-\frac{\tilde \beta_{t+1}}{2T}x_{t}+\sqrt{\frac{\tilde \beta_{t+1}}{T}}z_{t+1}.
$$
Then we rescale the time by a factor $1/T$, defining $t'=t/T$. With this we introduce $x(t')=x(t/T)=x_t$, $\beta(t')=\beta(t/T)=\tilde \beta_t$, $z(t')=z(t/T)=z_t$ and we denote $1/T$ by $\Delta t'$. 
%For $T\rightarrow \infty$, the new time $t'$ can be any element of the continuous interval $[0,1]$ and in this limit we denote $\Delta t'$ by $dt'$. 
Under these assumption we have that 
$$x(t'+\Delta t')-x(t')=-\frac{1}{2}\beta(t'+\Delta t')\Delta t' x(t') + \sqrt{\beta(t'+\Delta t')\Delta t'}z(t'+\Delta t').$$
For a big enough $T$, denoting $\Delta t'$ by $dt'$, we can heuristically say that the previous equation could be approximated as follows
\begin{equation}\label{stocdif}
\begin{split}
x(t'+\Delta t')-x(t')=dx(t')\approx -\frac{1}{2}\beta(t')x(t')dt' + \sqrt{\beta(t')}dw(t')
\end{split}
\end{equation}
where $dw(t)$ is the standard white noise and where we used the following approximations holding for $T$ big enough: since $\beta(t')<<1$, we can say that  $\beta(t'+dt)\approx \beta(t')$; $z(t'+ dt')\sqrt{dt'}\approx dw(t')$. The intuitive reason of the last approximation is that $z(t'+ \Delta t') \sim \mathcal{N}(0,I)$ and, with some abuse of notation, $dw_t \sim \mathcal{N}(0,dt)$. 

%Recall that the Markov process has transitions behaving like $p(x_{t}|x_{t-1})=\mathcal{N}(\sqrt{1-\beta_t}x_{t-1}; \beta_t I)$. So, heuristically, if we think to the process as 1-dimensional and in continuous time (for example with the transformation $t\mapsto t/T$ and $T\rightarrow \infty$), we can say that it behaves according to the stochastic differential equation
%\begin{equation}\label{SDE}
%dx_t=\sqrt{1-\beta_t}x_tdt+\beta_tdz_t,
%\end{equation}
%where $dz_t$ is the standard white noise. Note that, a similar equation can be assumed to hold in the case in which the $\{\beta_t\}_{t\in\{0,\dots,T\}}$ are really small. Indeed in this case $x_{t+1}$ is likely to be very close to $x_t$, and so it is reasonable to assume that both $x_t$ and $x_{t+1}$ belongs to a diffusion process. Moreover, as we saw previously, if $\beta_t<1$ for any $t \in \{0,\dots, T\}$ we are sure that the process admits a limiting distribution, and so that \eqref{SDE} admits a unique solution. 

The formulation \eqref{stocdif} for the dynamics of our model, thanks to the main theorem stated in \cite{reverse}, grants us that a reverse process exists and that it is a diffusion, as the forward process.  
\end{remark}

Now, that we have stated and proved the main properties of the forward process, we would like to find the best way to build the diffusion model which represents the reverse process starting from $x_T \sim \mathcal{N}(0,I)$ and going to some $x_0$ satisfying the properties of the starting real images  used for the forward process. Let us denote by $p_{\theta}(\cdot)$ the functions denoting the distributions (or conditional distributions) of the backward process depending on some set of unknown parameters $\theta$. The goal is finding the latent variables describing the limiting probability $p_{\theta}(x_0)=\int p_{\theta}(x_T,x_{T-1}, \dots, x_1, x_0) dx_{T}\dots dx_{1}$. Now, since the reverse of a Markov process still a Markov process (see \cite{Chung} for a detailed analysis of that), it is true that
\begin{equation}\label{markov2}
p_{\theta}(x_T,x_{T-1}, \dots, x_1, x_0)=p_{\theta}(x_T)\prod_{t=1}^Tp_{\theta}(x_{t-1}|x_t),
\end{equation}
which is the analogous of \eqref{markov1} but reverse in time. We already know that $p_{\theta}(x_{t-1}|x_{t})$ are Gaussian distributed (see Remark \ref{diffusion}) so, we can conclude that any of them is distributed like $\mathcal{N}(\mu_{\theta}(x_t), \Sigma_{\theta}(x_t))$ for some unknown $\mu_{\theta}(x_t)$ and $\Sigma_{\theta}(x_t)$. To evaluate \eqref{markov2} is then necessary to estimate $\mu_{\theta}(x_t)$ and $\Sigma_{\theta}(x_t)$ for each time $t \in \{1, \dots, T\}$. 

As pointed out in \cite{termo}, these optimal parameters are the ones that maximize the log likelihood averaged on the data that we have, which are given according to the distribution $p(x_0)$, being real images. This means that we have to find the $\mu_{\theta}(x_t)$ and $\Sigma_{\theta}(x_t)$ that maximize
\begin{equation}\label{loglike}
\mathcal{L}=\int \log(p_{\theta}(x_0)) p(x_0)dx_0.
\end{equation}
First, let us rewrite $p_{\theta}(x_0)$ as follows
\begin{equation}
\begin{split}
p_{\theta}(x_0)&=\int p_{\theta}(x_T,x_{T-1}, \dots, x_1, x_0) dx_{T}\dots dx_{1}\\&=\int \frac{p(x_0,\dots,x_T|x_0)}{p(x_0,\dots,x_T|x_0)}p_{\theta}(x_T)\prod_{t=1}^Tp_{\theta}(x_{t-1}|x_t)dx_{T}\dots dx_{1}\\&=\int p(x_1,\dots,x_T|x_0)p_{\theta}(x_T)\prod_{t=1}^T\frac{p_{\theta}(x_{t-1}|x_t)}{p(x_t|x_{t-1}, x_0)} dx_{T}\dots dx_{1}\\&=\mathbb{E}_{p(x_1,\dots,x_T|x_0)}\bigg[p_{\theta}(x_T)\prod_{t=1}^T\frac{p_{\theta}(x_{t-1}|x_t)}{p(x_t|x_{t-1}, x_0)}\bigg],
\end{split}
\end{equation}
where the second and third equality comes respectively from \eqref{markov2} and \eqref{markov1} plus the fact that $p(x_0,\dots,x_T|x_0)=p(x_1,\dots,x_T|x_0)$. Hence, by logarithm properties and Jensen's inequality, \eqref{loglike} can be bounded as follows

\begin{equation}\label{like}
\begin{split}
\mathcal{L}&=\int \log(p_{\theta}(x_0)) p(x_0)dx_0\\&=\int \log\Bigg(\mathbb{E}_{p(x_1,\dots,x_T|x_0)}\bigg[p_{\theta}(x_T)\prod_{t=1}^T\frac{p_{\theta}(x_{t-1}|x_t)}{p(x_t|x_{t-1}, x_0)}\bigg]\Bigg) p(x_0)dx_0\\
&
=\int \log\Bigg( \int p_{\theta}(x_T)\prod_{t=1}^T\frac{p_{\theta}(x_{t-1}|x_t)}{p(x_t|x_{t-1}, x_0)}p(x_1,\dots,x_T|x_0)dx_1\dots dx_T\Bigg) p(x_0)dx_0
\\
&
=\int \log\Bigg( \int p_{\theta}(x_T)\prod_{t=1}^T\frac{p_{\theta}(x_{t-1}|x_t)}{p(x_t|x_{t-1}, x_0)}d\tilde p(x_1,\dots,x_T|x_0)\Bigg) p(x_0)dx_0
\\
& \geq\int \Bigg(\int \log\Big( p_{\theta}(x_T)\prod_{t=1}^T\frac{p_{\theta}(x_{t-1}|x_t)}{p(x_t|x_{t-1}, x_0)} \Big)d\tilde p(x_1,\dots,x_T)\Bigg)p(x_0)dx_0 \\
& = \int \log\Big( p_{\theta}(x_T)\prod_{t=1}^T\frac{p_{\theta}(x_{t-1}|x_t)}{p(x_t|x_{t-1}, x_0)} \Big) p(x_1,\dots,x_T|x_0)p(x_0)dx_0dx_1\dots dx_T\\
& = \int \log\Big( p_{\theta}(x_T)\prod_{t=1}^T\frac{p_{\theta}(x_{t-1}|x_t)}{p(x_t|x_{t-1}, x_0)} \Big) p(x_0,x_1,\dots,x_T)dx_0dx_1\dots dx_T\\
& = \mathbb{E}_{p(x_0,\dots,x_T)}\bigg[ \log\Big( p_{\theta}(x_T)\prod_{t=1}^T\frac{p_{\theta}(x_{t-1}|x_t)}{p(x_t|x_{t-1}, x_0)} \Big) \bigg]\\
& = \mathbb{E}_{p(x_T)}\big[\log(p_{\theta}(x_T))\big]+\mathbb{E}_{p(x_0,\dots,x_T)}\bigg[ \log\Big( \prod_{t=1}^T\frac{p_{\theta}(x_{t-1}|x_t)}{p(x_t|x_{t-1}, x_0)} \Big) \bigg],
\end{split}
\end{equation}
where used the notation $d\tilde p(x_1,\dots,x_T|x_0)$ to denote the random probability measure of density $p(x_1,\dots,x_T|x_0)$.
Observe that the first term of the last bound above does not depend on the parameters $\theta$ since we showed that $x_T$ is distributed as a $\mathcal{N}(0,I)$. 
Then the parameters that maximize $\mathcal{L}$ are the same that maximize
$$
\mathbb{E}_{p(x_0,\dots,x_T)}\bigg[ \log\Big( \prod_{t=1}^T\frac{p_{\theta}(x_{t-1}|x_t)}{p(x_t|x_{t-1}, x_0)} \Big) \bigg].
$$
 
 Moreover, let us observe that the first factor of this term ($t=1$) is equal to
\begin{equation}\label{marginal0}
\begin{split}
\mathbb{E}_{p(x_0,x_1)}\bigg[ \log\Big(\frac{p_{\theta}(x_{0}|x_1)}{p(x_1| x_0)} \Big) \bigg]&=\mathbb{E}_{p(x_0,x_1)}\big[ \log(p_{\theta}(x_{0}|x_1)) \big]-\mathbb{E}_{p(x_0,x_1)}\big[ \log(p(x_{1}|x_0)) \big].
\end{split}
\end{equation}
%indeed the second term on the right hand-side of the the first equality is $0$: under the distribution $p(x_0)$, $p(x_1|x_0)=1$ (intuitively, given our data, which comprehends $x_0$ and $x_1$ we are sure that the process goes from $x_0$ to $x_1$, so $p(x_1|x_0)=1$). 
%Therefore, summarizing, maximize $\mathcal{L}$ is equivalent to maximize 
%\begin{equation}
%K:=\mathbb{E}_{p(x_0,x_1)}\big[ \log(p_{\theta}(x_{0}|x_1)) %\big]+ \mathbb{E}_{p(x_0,\dots,x_T)}\bigg[ \log\Big( %\prod_{t=2}^T\frac{p_{\theta}(x_{t-1}|x_t)}{p(x_t|x_{t-1}, %x_0)} \Big) \bigg].
%\end{equation}
Now, thanks to Bayes Theorem we can write $$p(x_t|x_{t-1},x_0)=\frac{p(x_{t-1}|x_t,x_0)p(x_t|x_0)}{p(x_{t-1}|x_0)}$$ and, with this in mind, we rewrite all the factors of $\tilde{\mathcal{L}}$ but the first (so all the ones involving $t\geq 2$) as 
\begin{equation}\label{center}
\begin{split}
\sum_{t=2}^T&\int  \log\bigg(\frac{p_{\theta}(x_{t-1}|x_t)}{p(x_t|x_{t-1}, x_0)} \bigg)p(x_0,\dots,x_T)dx_0,\dots dx_T \\ &= \sum_{t=2}^T\int  \log\bigg(\frac{p_{\theta}(x_{t-1}|x_t)}{p(x_{t-1}|x_t,x_0)}\frac{p(x_{t-1}|x_0)}{p(x_t|x_0)} \bigg)p(x_0,\dots,x_T)dx_0,\dots dx_T \\&=\sum_{t=2}^T\int  \log\bigg(\frac{p_{\theta}(x_{t-1}|x_t)}{p(x_{t-1}|x_t,x_0)} \bigg)p(x_0,\dots,x_T)dx_0,\dots dx_T\\& \qquad \qquad +\sum_{t=2}^T\int  \log\bigg(\frac{p(x_{t-1}|x_0)}{p(x_t|x_0)} \bigg)p(x_0,\dots,x_T)dx_0,\dots dx_T.
\end{split}
\end{equation}
%Observe now that the second term of the last sum obtained above does not depend on the parameters we want to estimate, so it is not useful for our study. 
Let us focus first on the first term on the right hand-side of the last equality obtained above. Observe that, since each $\log\bigg(\frac{p_{\theta}(x_{t-1}|x_t)}{p(x_{t-1}|x_t,x_0)} \bigg)$ depends only on $x_0,x_{t-1}, x_t$, we can rewrite
\begin{equation}
\begin{split}
\sum_{t=2}^T&\int  \log\bigg(\frac{p_{\theta}(x_{t-1}|x_t)}{p(x_{t-1}|x_t,x_0)} \bigg)p(x_0,\dots,x_T)dx_0,\dots dx_T
\\& =\sum_{t=2}^T\int  \log\bigg(\frac{p_{\theta}(x_{t-1}|x_t)}{p(x_{t-1}|x_t,x_0)} \bigg)p(x_0,x_{t-1},x_t)dx_0,dx_{t-1}, dx_t
\\& =\sum_{t=2}^T\int  \log\bigg(\frac{p_{\theta}(x_{t-1}|x_t)}{p(x_{t-1}|x_t,x_0)} \bigg)p(x_{t-1}|x_t,x_0)p(x_t,x_0)dx_0 dx_{t-1}dx_t\\&=-\sum_{t=2}^T\int  \log\bigg(\frac{p(x_{t-1}|x_t,x_0)}{p_{\theta}(x_{t-1}|x_t)} \bigg)p(x_{t-1}|x_t,x_0)p(x_t,x_0)dx_0 dx_{t-1}dx_t\\&=-\sum_{t=2}^T\mathbb{E}_{p(x_0,x_t)}\Big[D_{KL}\big(p(x_{t-1}|x_t,x_0) \;||\; p_{\theta}(x_{t-1}|x_t)\big)\Big];
\end{split}
\end{equation}
where $D_{KL}$ denotes the Kullback–Leibler divergence (which is a function of $x_t$ and $x_0$ in this case), i.e. for any $x_t,x_0$,
$$D_{KL}\big(p(x_{t-1}|x_t,x_0) \;||\; p_{\theta}(x_{t-1}|x_t)\big):=  \int  \log\bigg(\frac{p(x_{t-1}|x_t,x_0)}{p_{\theta}(x_{t-1}|x_t)} \bigg)p(x_{t-1}|x_t,x_0) dx_{t-1}.$$

Now, we analyze the second term on the right hand-side of the last equality in \eqref{center}. Using the properties of logarithm it can be written as
\begin{equation}
\begin{split}       
\sum_{t=2}^T&\int \Big( \log(p(x_{t-1}|x_0))-\log({p(x_t|x_0)} )\Big)p(x_0,x_{t-1},x_t)dx_0, dx_{t-1}, dx_t\\
&=\sum_{t=2}^T\Big( \mathbb{E}_{p(x_{t-1}, x_0)}\big[\log(p(x_{t-1}|x_0))\big]-\mathbb{E}_{p(x_{t}, x_0)}\big[\log(p(x_{t}|x_0))\big]\Big)\\
&=\mathbb{E}_{p(x_{1}, x_0)}\big[\log(p(x_{1}|x_0))\big]-\mathbb{E}_{p(x_{T}, x_0)}\big[\log(p(x_{T}|x_0))\big],
\end{split}
\end{equation}
where the last equality is due to the telescopic property of the sum obtained before. Observe now that the first term on the right hand-side of the last equality above cancels with the second on the right hand-side of \eqref{marginal0}. At the same time we can write the sum of the second term on the right hand-side of the last equality above plus the first on the right hand-side of the last equality in \eqref{like} as
\begin{equation}
\begin{split}
\mathbb{E}_{p(x_T)}&\big[\log(p_{\theta}(x_T))\big]-\mathbb{E}_{p(x_{T}, x_0)}\big[\log(p(x_{T}|x_0))\big]\\&=\int \log \bigg(\frac{p_{\theta}(x_T))}{p(x_{T}|x_0)}\Big) p(x_0,x_T)dx_0dx_T\\&=\int \log \bigg(\frac{p_{\theta}(x_T))}{p(x_{T}|x_0)}\Big) p(x_T|x_0)p(x_0)dx_0dx_T\\&=-\int \log \bigg(\frac{p(x_{T}|x_0)}{p_{\theta}(x_T))}\Big) p(x_T|x_0)p(x_0)dx_0dx_T\\&=-\mathbb{E}_{p(x_0)}\Big[D_{KL}\big(p(x_{T}|x_0)\;||\; p_{\theta}(x_T) \big) \Big].
\end{split}
\end{equation}

Finally, summing all the term remained, we can conclude that 
\begin{equation}\label{tomax1}
\begin{split}
\mathcal{L}=\mathbb{E}_{p(x_0,x_1)}\big[\log(p_{\theta}(x_{0}|x_1))\big]-&\sum_{t=2}^T\mathbb{E}_{p(x_0,x_t)}\Big[D_{KL}\big(p(x_{t-1}|x_t,x_0) \;||\; p_{\theta}(x_{t-1}|x_t)\big)\Big]\\&-\mathbb{E}_{p(x_0)}\Big[D_{KL}\big(p(x_{T}|x_0)\;||\; p_{\theta}(x_T) \big)\Big].
\end{split}
\end{equation}
Since we proved that $p_{\theta}(x_T)$ under the opportune conditions on the parameters $\{\beta_t\}_{t=1}^T$ is a $\mathcal{N}(0,I)$, it does not depends on the parameters $\theta$. So, maximising \eqref{tomax1} is equivalent to maximise 
\begin{equation}\label{tomax}
\begin{split}
\tilde{\mathcal{L}}:=\mathbb{E}_{p(x_0,x_1)}\big[\log(p_{\theta}(x_{0}|x_1))\big]-\sum_{t=2}^T\mathbb{E}_{p(x_0,x_t)}\Big[D_{KL}\big(p(x_{t-1}|x_t,x_0) \;||\; p_{\theta}(x_{t-1}|x_t)\big)\Big],
\end{split}
\end{equation}
 Hence, the model is given by the solutions of the problem $\hat p_{\theta}(x_{t-1}|x_{t})=\argmax_{p_{\theta}(x_{t-1}|x_{t})} \tilde{\mathcal{L}}$. 

\begin{remark}
It is possible to evaluate explicitly the value of $p(x_{t-1}|x_t,x_0)$ as it is mentioned in \cite{denoising} and explained in detail in the next Section.
\end{remark}

\subsubsection{Analytic computation of $p(x_{t-1}|x_t,x_0)$}
We know that $p(x_{t-1}|x_t,x_0)$ is Gaussian distributed as a $\mathcal{N}(\tilde{\mu}(x_t,x_0),  \tilde{\Sigma}(x_t,x_0))$. In \cite{denoising} the values of $\tilde \mu(x_t,x_0)$ and $\tilde \Sigma(x_t,x_0))$ are given and they are useful for the estimate procedure. In this brief Section we give a strategy of evaluating these values using the results of \cite{bishop}, Chapter 2.3.3 plus some computations. 

For that purpose we use the same notation used in Chapter 2.3.3 of \cite{bishop}, i.e. we denote $p(x_{t-1}|x_t,x_0)$ by $p(x|y)$, $p(x_{t}|x_{t-1},x_0)=p(x_{t}|x_{t-1})$ by $p(y|x)$ and $p(x_{t-1}|x_0)$ by $p(x)$. By our previous computations we already know that $p(y|x)$ and $p(x)$ are respectively distributed like two Gaussian variables $\mathcal{N}(\sqrt{1-\beta_t}x_{t-1}, \beta_t I)$ and $\mathcal{N}(\sqrt{\overline \alpha_{t-1}}x_0, (1-\overline{\alpha}_{t-1})I)$. Now, for simplify the notation, let us consider the unidimensional case, the general case is totally analogous being the covariance matrices diagonal.

So, keeping in mind the relation $
(1-\beta_t)(1-\overline \alpha_{t-1})=1-\beta_t-\overline\alpha_t
$ and applying (2.116) of \cite{bishop} to our setting we get that:
\begin{equation}\label{var}
\begin{split}
\tilde{\Sigma}(x_t,x_0)=\tilde{\Sigma}_t=\bigg(\frac{1}{1-\overline \alpha_{t-1}}+(1-\beta_t)\frac{1}{\beta_t}\bigg)^{-1}&=\bigg(\frac{\beta_t+(1-\beta_t)(1-\overline \alpha_{t-1})}{\beta_t(1-\overline \alpha_{t-1})}\bigg)^{-1}\\&=\beta_t \frac{(1-\overline \alpha_{t-1})}{(1-\overline \alpha_{t})}
\end{split}
\end{equation}
and
\begin{equation}\label{mean}
\begin{split}
\tilde{\mu}(x_t,x_0)&=\tilde{\Sigma}_t\bigg(\sqrt{1-\beta_t}\frac{1}{\beta_t}x_t+\frac{1}{1-\overline \alpha_{t-1}}\sqrt{\overline \alpha_{t-1}}x_0 \bigg)\\&=\frac{\sqrt{\alpha_t}(1-\overline\alpha_{t-1})}{1-\overline\alpha_t}x_t+\frac{\beta_t \sqrt{\overline\alpha_{t-1}}}{1-\overline \alpha_{t}}x_0,
\end{split}
\end{equation}
as stated in \cite{denoising}.

\subsection{Estimate procedure}
In this section we want to analyze further the problem of maximization of \eqref{tomax}, making assumptions on the distributions $p_{\theta}(x_{t-1}|x_t)$.
We already saw that $p_{\theta}(x_{t-1}|x_t)$ is distributed like a $\mathcal{N}(\mu_{\theta}(x_t), \Sigma_{\theta}(x_t))$. Proceeding as it is done in \cite{denoising}, we assume that, for any $t \in \{1, \dots, T \}$,  $\Sigma_{\theta}(x_t)=\sigma^2_t I$ for some experimentally established value of $\sigma_t^2$ and so, it does not depend on the process itself nor on the parameters $\theta$ (this is fundamental for the derivation of \eqref{KLprop}). In \cite{denoising} the authors suggested two possibilities: $\sigma^2_t=\beta_t$ or $\sigma^2=\beta_t\frac{1-\overline{\alpha}_{t-1}}{1-\overline{\alpha}_{t}}$ .

Under these assumptions, thanks to the properties of Kullback–Leibler divergence of two multivariate normal distribution it is true that 

\begin{equation}\label{KLprop}
D_{KL}\big(p(x_{t-1}|x_t,x_0) \;||\; p_{\theta}(x_{t-1}|x_t)\big)=\frac{1}{2\sigma^2_t}||\tilde \mu(x_t,x_0)-\mu_{\theta}(x_t)||^2 + C,
\end{equation}
where $C$ is some constant not depending on $\theta$ and $||\cdot||$ is the standard $L^2$ norm. Moreover, recalling that we can express $x_t$ as a function of $x_0$ as 
\begin{equation}\label{17}
x_t(x_0)=\sqrt{\overline \alpha_t} x_0 + \sqrt{1-\overline \alpha_t}z_t
\end{equation}
for some $z_t \sim \mathcal{N}(0,I)$, we can rewrite the second term on the right had-side of \eqref{tomax} as
\begin{equation}\label{exp1}
\begin{split}
\sum_{t=2}^T&\mathbb{E}_{p(x_0)}\Big[\frac{1}{2\sigma^2_t}||\tilde \mu(x_t(x_0),x_0)-\mu_{\theta}(x_t(x_0))||^2\Big]+C'\\ &=\sum_{t=2}^T\mathbb{E}_{p(x_0), z_t}\Big[\frac{1}{2\sigma^2_t}\Big|\Big|\tilde \mu\Big(x_t(x_0),\frac{1}{\sqrt{\overline\alpha_t}}(x_t(x_0)-\sqrt{1-\overline \alpha_t}z_t)\Big)-\mu_{\theta}(x_t(x_0))\Big|\Big|^2\Big]+C',
\end{split}
\end{equation}
where $C'$ is a constant not depending on $\theta$ and the $z_t$ at pedix of the second expectation means that we have to average also on the different possible values that $z_t \sim \mathcal{N}(0,1)$ takes at any time step. Now, thanks to \eqref{mean} we can rewrite the argument of the expectation in the last term above as 
\begin{equation}\label{exp}
\begin{split}
&\frac{1}{2\sigma^2_t}\Big|\Big|\frac{\sqrt{\alpha_t}(1-\overline{\alpha}_{t-1})}{1-\overline\alpha_t}x_t(x_0)+\frac{\beta_t\sqrt{\overline{\alpha}_{t-1}}}{1-\overline\alpha_t}\frac{1}{\sqrt{\overline{\alpha}_t}}(x_t(x_0)-\sqrt{1-\overline{\alpha}_t}z_t)-\mu_{\theta}(x_t(x_0))\Big|\Big|^2\\
&=\frac{1}{2\sigma^2_t}\Big|\Big|\frac{{\alpha_t}(1-\overline{\alpha}_{t-1})+(1-\alpha_t)}{\sqrt{\alpha_t}(1-\overline\alpha_t)}x_t(x_0)-\frac{\beta_t}{\sqrt{1-\overline{\alpha}_t}}\frac{1}{\sqrt{{\alpha}_t}}z_t-\mu_{\theta}(x_t(x_0))\Big|\Big|^2\\
&=\frac{1}{2\sigma^2_t}\Big|\Big|\frac{1}{\sqrt{\alpha_t}}\Big(x_t(x_0)-\frac{\beta_t}{\sqrt{1-\overline{\alpha}_t}}z_t\Big)-\mu_{\theta}(x_t(x_0))\Big|\Big|^2.
\end{split}
\end{equation}
From these computation it becomes clear that the best $\mu_{\theta}(x_t(x_0))$ is the one which predicts $\frac{1}{\sqrt{\alpha_t}}\big(x_t(x_0)-\frac{\beta_t}{\sqrt{1-\overline{\alpha}_t}}z_t\big)$. Moreover, observe that $x_t$ is actually an input of the model and so we could take (as suggested in \cite{denoising}):
\begin{equation}\label{ztheta}
\begin{split}
\mu_{\theta}(x_t(x_0))&=\tilde \mu\big(x_t(x_0),\frac{1}{\sqrt{\overline\alpha_t}}(x_t(x_0)-\sqrt{1-\overline \alpha_t}z_{\theta}(x_t))\big)\\&=\frac{1}{\sqrt{\alpha_t}}\Big(x_t(x_0)-\frac{\beta_t}{\sqrt{1-\overline{\alpha}_t}}z_{\theta}(x_t)\Big),
\end{split}
\end{equation}
where $z_{\theta}(x_t)$ is the function predicting $z_t$ from the value of the input $x_t$ and the equality comes from \eqref{mean} plus some simple computations.

Hence, thanks to \eqref{exp} and \eqref{ztheta}, we get that the expectation in \eqref{exp1} is equal to
\begin{equation}
\begin{split}
\mathbb{E}_{p(x_0), z_t}\Big[&\frac{\beta_t^2}{2\sigma^2_t\alpha_t(1-\overline \alpha_t)}\big|\big|z_t-z_{\theta}(x_t)\big|\big|^2\Big]\\
&=\mathbb{E}_{p(x_0), z_t}\Big[\frac{\beta_t^2}{2\sigma^2_t\alpha_t(1-\overline \alpha_t)}\big|\big|z_t-z_{\theta}(\sqrt{\overline \alpha_t} x_0 + \sqrt{1-\overline \alpha_t}z_t)\big|\big|^2\Big],
\end{split}
\end{equation} 
where the equality comes from \eqref{17}. The function that we are looking for is therefore the $z_{\theta}(\cdot)$ that minimize the quantity 
\begin{equation}\label{likelihoodmax}
\sum_{t=2}^T\mathbb{E}_{p(x_0), z}\Big[\frac{\beta_t^2}{2\sigma^2_t\alpha_t(1-\overline \alpha_t)}\big|\big|z_t-z_{\theta}(\sqrt{\overline \alpha_t} x_0 + \sqrt{1-\overline \alpha_t}z_t)\big|\big|^2\Big].
\end{equation}

\begin{remark}
	In \cite{denoising} the authors empirically showed that in order to get the function $z_{\theta}$ it is possible to minimize the following quantity instead of \eqref{likelihoodmax} and what we define in Section \ref{t=1}:
	\begin{equation}
	\sum_{t=2}^T\mathbb{E}_{p(x_0), z_t}\Big[\big|\big|z_t-z_{\theta}(\sqrt{\overline \alpha_t} x_0 + \sqrt{1-\overline \alpha_t}z_t)\big|\big|^2\Big].
	\end{equation}
	Actually they showed that minimizes this unweighted functional gives even better performance.
\end{remark}

Notice that, thanks to the assumption that we made at the beginning of this Section, once we have estimated the function $z_{\theta}(\cdot)$ we ``have'' the reverse process. Indeed, since $p(x_{t-1}|x_t)=\mathcal{N}\bigg(\frac{1}{\sqrt{\alpha_t}}\Big(x_t(x_0)-\frac{\beta_t}{\sqrt{1-\overline{\alpha}_t}}z_{\theta}\Big),\sigma^2_t I\bigg)$, we have that 
\begin{equation}\label{rev_eq}
x_{t-1}=\frac{1}{\sqrt{\alpha_t}}\Big(x_t-\frac{\beta_t}{\sqrt{1-\overline{\alpha}_t}}z_{\theta}(x_t)\Big)+\sigma_t z_t.
\end{equation}

\begin{remark}
The last computations (after \eqref{KLprop}) are important to make comprehensible that what we are doing in the whole estimation process is, in some way, to estimate the noise to remove at each step in the reverse process. More in practice they are not necessary, in principle it is indeed possible simply to build some model able to find directly $\mu_{\theta}$ maximizing the left hand-side of \eqref{exp1}. Then the estimated reverse process would simply be given by 
\begin{equation}\label{rev_eq2}
x_{t-1}=\mu_{\theta}(x_t)+\sigma_t z_t,
\end{equation}
equivalent to \eqref{rev_eq} before all the computations showed.
\end{remark}

\begin{figure}[ht]
  \includegraphics[width=\linewidth, trim={0cm 0cm 0cm 0cm}]{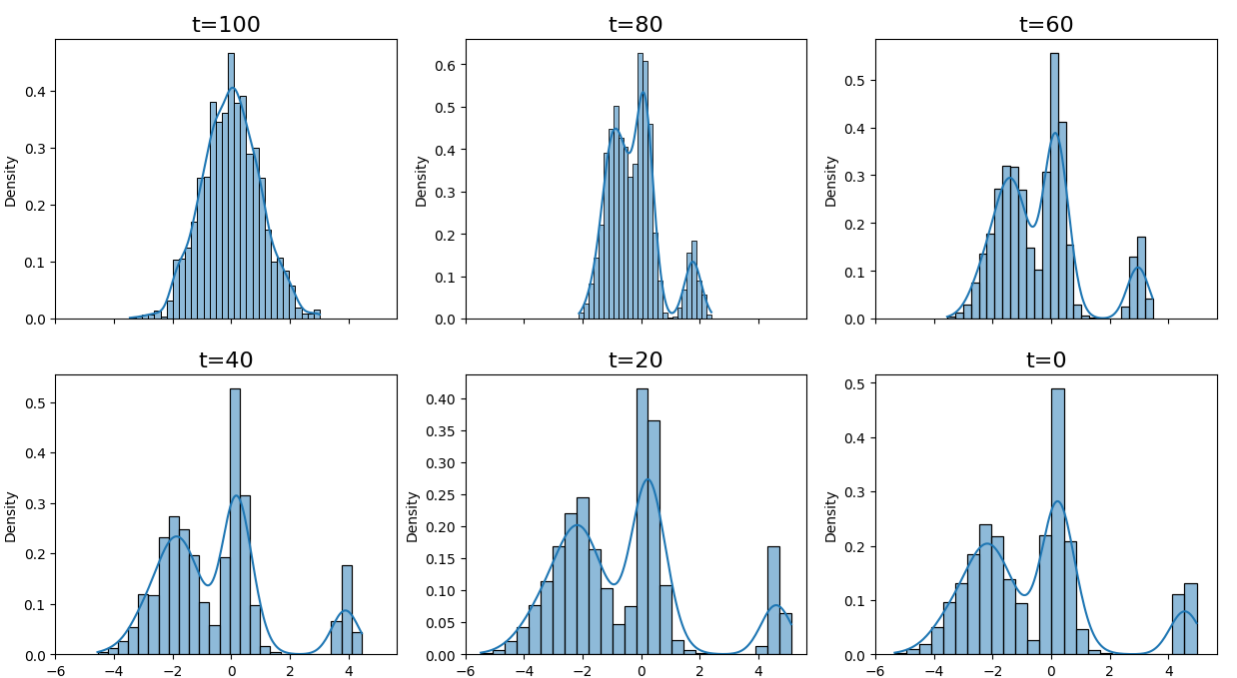}
  \caption{Distributions obtained with the reverse process obtained with \eqref{rev_eq} starting from $2000$ samples from a normal distribution $\mathcal{N}(0,1)$. For the training of the model were used the trajectories simulated for getting the distributions in Figure \ref{fig:example}, indeed it is clear as the limiting distribution of the reverse process in this picture is close to the initial distribution of data there.}
  \label{fig:example}
\end{figure}

%\begin{remark}
%The fact that the reverse process is a diffusion is the key to arrive to this result and at the same time it is the cause for which there is the term $\sigma_t z$ in \eqref{rev_eq}. The coefficient $\sigma$ is arbitrary thanks to the assumptions that we made and in \cite{denoising} two possibilities are presented  (both of them small values)  with their pro and cons.
%\end{remark}

\subsubsection{Analysis of $\mathbb{E}_{p(x_0,x_1)}\big[\log(p_{\theta}(x_{0}|x_1))\big]$}\label{t=1}
The last step of the reverse process determined by $p_{\theta}(x_{0}|x_1)$ is treated in section 3.3 of \cite{denoising} in a particular way. Indeed, the authors proposed to work through the model with images which are combinations of pixel values (usually in the set $\{0,1,\dots,255\}$) rescaled to combinations of element in the interval $[-1,1]$. Then within the forward process Gaussian noise is added to the ``scaled''  images up to the limit distributions which, theoretically, lives in the space of combination of elements in $[-\infty, +\infty]$. In particular, starting from a normal distributed noisy image $x_T$, in principle the reverse process built goes up to an image $x_0$ which still lives in the space of combination of elements in $[-\infty, +\infty]$. For this reason the authors, in order of having as an output of the reverse process an image in the original pixel space (which can be deducted from the space of combination of elements in $[-1.1]$), define the last step $p_{\theta}(x_0|x_1)$ to be a discrete decoder depending on $\mu_{\theta}(x_1(x_0))$ in such a way that, receiving in input an image $x_1$ in the unlimited space, it returns as output an image in the scaled space $x_0$ which can then be converted to the pixel space.

This is a technicality necessary for the application of this kind of model but does not change the mathematical theory behind DPM which we presented in the previous sections. So, we refer the interested reader to the aforementioned section of \cite{denoising} in which it is also pointed out that the solution proposed is not unique and neither optimal.

\section{DPM on latent space - Stable Diffusion}\label{stable_sec}
The most important mathematical theory for DPM is the one presented in Section \ref{diffusion_paper} which was developed first in \cite{termo} and which is behind the outstanding results of \cite{denoising}. This kind of model is built in the pixel space. Indeed, for any $t \in \{0,\dots, T\}$ the values taken by the forward and backward process $x_{t}$ are $H \times W \times 3$ matrices, where $H$ and $W$ are respectively the height and the width of the images in pixels and $3$ are the RGB values for each pixel (eventually rescaled, see Section \ref{t=1}).  This feature translates in the fact that training, using and optimizing such models is computationally really expansive and therefore, slow.

In \cite{stable} the authors proposed some important changes to the DPM that, using the same mathematical background that we explained in Section \ref{background}, reduce the dimension of the space in which the model is built, improving its performance. The main idea is to create a model which is not working anymore on the pixel space but on the space of some kind of compressed images. So, intuitively they do not train a model that is able to denoise images, but they train a model that is able to denoise compressed representations of them. In such a way, when the trained model generates a new compressed image, starting from the compressed representation of noise, it is possible to retrieve with some decoder the image corresponding to that representation. This leads to an huge advantage in term of computation complexity. This kind of model is usually called stable diffusion (DS hereinafter).

In the next sections we present the principal ideas behind DS, without entering in the detail of the neural architecture used to implement it. It is indeed well explained in the original paper \cite{stable} and it goes beyond the goal of this work that aims meanly to give an intuitive idea of the theory behind generative diffusion models.

\subsection{From pixel to latent space}
As we anticipated, in the model presented in \cite{stable} the authors propose to change the space in which the diffusion model is built, passing from the images described pixel per pixel, so in the space $\Gamma=C^{H\times W\times 3}$ where $C$ depends on the number of color considered (usually it is the space $\{0,\dots, 255\}$), to representations of these images in some other smaller space $\tilde \Gamma=C^{h\times w\times 3}$, with $h$ and $w$ such that $H/h=W/w=2^m$ for some $m \in \mathbb{N}$. In their paper this is done using a perceptrual compression model based on the work \cite{decoder} that, given an image $x \in \Gamma$, performs the transformation $y=\mathcal{E}(x) \in \tilde \Gamma$, where $\mathcal{E}$ represents the encoder. We denote by $\mathcal{D}$ the decoder which is able to transform any element of the latent space $\tilde y \in \tilde \Gamma$ back to an image in the pixel space, i.e. $x' = \mathcal{D}(\tilde y)= \mathcal{D}(\mathcal{E}(\tilde x)) \in \Gamma$ for some $\tilde x \in \Gamma$.

The basic idea behind this kind of image compression is that it permits to get the ``essence'' of an image in a smaller representation, in such a way that all the important features of the original image are, in some way, preserved and represented in a smaller space. This does not only give an important advantage in terms of computational cost but also permits to train the model only on the parts that really ``mean something'' of the images simply minimizing the function (equivalent to \eqref{likelihoodmax} for the DPM)
	\begin{equation}
	\sum_{t=2}^T\mathbb{E}_{y_t=\mathcal{E}(x_t), \tilde z_t}\Big[\big|\big|\tilde z_t-\tilde z_{\theta}(y_t)\big|\big|^2\Big],
	\end{equation}
 where the $x_t$ are distributed according to $p(x_t)$ for any $t \in \{0,\dots, T\}$, $\tilde z_t$ is the noise in the latent space and $\tilde z_{\theta}$ is the estimate for the noise $\tilde z_t$ in the latent space. So, the whole model moved  in the latent space, but it is not a problem, indeed when we recover the limit for the reverse process in this space it can also be transformed back in the pixel space. Indeed, when the model is trained it is able to recover, from a latent variable $y_T=\mathcal{E}(x_T)$ for some $x_T \sim \mathcal{N}(0,I)$, a limit distribution $y_0$ which is the representation of some realistic image $x_0=\mathcal{D}(y_0)$ in the pixel space $\Gamma$. 

 \subsection{Conditioning}\label{condit}
 Working on a latent abstract space not only makes the model lighter to train and use, but, as pointed out in \cite{stable} it also allows to add the possibility of conditioning the image generation process on some additional input given by the user via a prompt. It is indeed possible to train the model conditioned on some (a lot of) inputs given by the user, training it on a larger set of possible images but associated to some representation of the user input. 
 
 It is easier to understand making a toy example: we can train the model passing a lot of images representing cats and dogs associated, respectively, to the user input ``cat'' and ``dog''. All of these, images and inputs, encoded in some latent variable space. Once it is trained on the representation of images (to which noise is added) and inputs, the model will generate outputs depending on the prompt inserted. So that, if we ask for a ``cat'' the whole reverse process will be conditioned to converge to some latent representation of realistic image of a cat. If the model is trained on a lot of possible users inputs and a lot of images for each of them, it will translate any input received in a combination of the ones on which it is trained and then condition the reverse process to converge to some representation of an image which is associated to this combination of inputs.
 
 Hence, these inputs have to be seen as additional variables on which the function $\tilde z_{\theta}$ depends, i.e. before this function was only dependent on time and the latent variable $y_t$, now we have to add the dependence on some encoded input $\iota$. So, the new function to minimize is 
 \begin{equation}
	\sum_{t=2}^T\mathbb{E}_{y_t=\mathcal{E}(x_t), \tilde z, \iota}\Big[\big|\big|\tilde z-\tilde z_{\theta}(y_t, \iota)\big|\big|^2\Big],
\end{equation}
where $\iota$ is taken on the space of all the encoded inputs for which the model is trained. 

\begin{figure}[ht]
  \includegraphics[width=\linewidth, trim={0cm 8cm 14cm 0.5cm}]{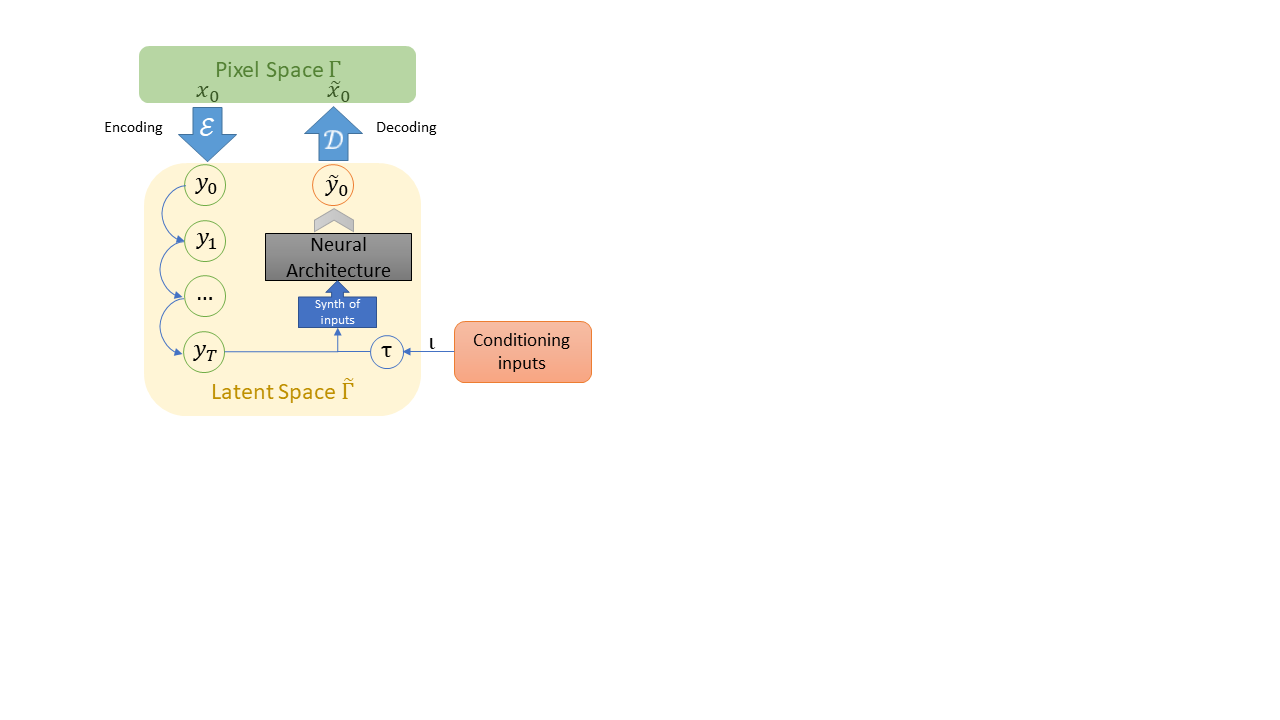}
  \caption{Scheme of what we described in Section \ref{condit}. On the left is represented the process in which a pixel image $x_0$ is encoded to a vector $y_0$ in the latent space. This vector is then transformed, adding noise at each step (diffusion process), in a noised vector $y_T$. The reverse process starts synthetizing the conditioning input to the noised image and then passing everything to the principal neural architecture which gives back a denoised $\tilde y_0$ (conditioned on $\iota$) that is finally decoded to a (new) pixel image $\tilde x_0$.}
  \label{fig:stable_scheme}
\end{figure}

The equation before is not exactly what the neural architecture does, it is much more complex because it is necessary to include, in some way, any possible input $\iota$. The authors in \cite{stable} to reach that, defined a projection $\tau$ from the space of all possible $\iota$ to a latent space which can be integrated in to the model being (in some sense) a combination of the inputs that the model received during training. 

\begin{remark}
    It is clear that a model such that need to be trained on a huge set of combinations of images and user inputs, bigger than the one used for DPM that is usually focused on some specific type of images (for example celebrities, cats, churches...). It is possible indeed to see the DS as a combination of many DPM, each of them associated to some kind of input given by the user. Then, once the user ask for some specific kind of image, the DS model conditions itself to generate that kind of image, that could be generated also by a DPM trained on a set of images corresponding to that specific input.
\end{remark}

\section{Cold Diffusion: a deterministic generative model}\label{cold_sec}
In the recent paper \cite{cold} the authors found out that all the complex theory behind DPM and DS can be noticeably simplified without considering the forward process (that add noise) as a random process, but taking a completely deterministc one.

Indeed, they show empirically that similar models can be built using arbitrary degradation processes and not only the one used for the DPM that add Gaussian noise at each step. Below we formalize this concept as it is done in \cite{cold}.

Let us define the following degradation operator:
\begin{equation}
\begin{split}
D:\Gamma\times\{0,\dots, T\} &\longrightarrow \Gamma\\
(x,t) &\longmapsto D(x,t),
\end{split}
\end{equation}
that at each image $x \in \Gamma$ associate its degradation $D(x,t)$ after $t$ degradation steps. So, given a starting image $x_0$, $x_t=D(x_0, t)$ is the analogous of \eqref{17} in the DPM, but with an arbitrary degradation transformation $D$. 

\begin{remark}
    Considering the (random) operator $D$ defined on any $(x_0,t)\in\Gamma\times\{0,\dots, T\}$ by $D(x_0, t)=\sqrt{\overline \alpha_t} x_0 + \sqrt{1-\overline \alpha_t}z_t$ we retrieve exactly the DPM.
\end{remark}

Now, it is necessary to define the equivalent of the reverse process in DPM and DS. In \cite{cold} it is done defining an operator that restores the corrupted images, which plays the role of the denoising process developed in the case of the DPM. This operator is defined as
\begin{equation}
\begin{split}
R:\Gamma\times\{0,\dots, T\} &\longrightarrow \Gamma\\
(x,t) &\longmapsto R(x,t),
\end{split}
\end{equation}
such that, given a forward process $x_0, x_1, \dots, x_t$, $R(x_t,t)$ is an approximation of $x_0$. 

The operator $R$ depends on some unknown parameters $\theta$ and so we denote it by $R=R_{\theta}$. The goal of the neural model is to estimate in some way this operator, so that it could be able to restore, from some corrupted image $x_T$, a realistic (and new) image $\tilde x_0$, which as we said is an approximation (but not the same) of the image $x_0$ such that $x_T=D(x_0,t)$. This is the analogous of estimating the function $z_{\theta}$ in order to denoise the image $x_T\sim \mathcal{N}(0,I)$ in the DPM.

The loss function to be minimized in order to find the operator $R_{\theta}$ proposed in \cite{cold} is simply
\begin{equation}
\mathbb{E}_{x \sim \mathcal{X}, t}\big[||R_{\theta}(D(x,t),t)-x||\big]
\end{equation}
where $\mathcal{X}$ is the set of (not corrupted) images that we use in training and $||\cdot||$ is the $\ell_1$ norm, i.e. for any $N$ and $a=(a_{1},\dots, a_N) \in \mathbb{R}^N$, $||a||=\sum_{i=1}^N|a_i|$.

\begin{remark}
Notice an important difference between this kind of model and the DPM: in this case indeed the output of the model is an operator which, in some sense, restores the image in one step, i.e. if we have $x_t$ which was corrupted $t$ times via the operator $D$, we can recover immediately $x_0=R(x_t,t)$; while in DPM this was done denoising at each step the noise image, going from $x_T$ to $x_{T-1}$, then to $x_{T-2}$ and so on up to get to an $x_0$ corresponding to a real image. In \cite{cold} the authors proved empirically that restoring the image in one step as $x_0=R(x_t,t)$ performed badly compared to combine $R$ in sequence with $D$ to get some sort of step by step restoration as it is done in Algorithm \ref{alg}. Actually in \cite{cold} the authors also presented a better algorithm to restoring images, but it goes beyond the goal of this paper and so we refer the interested reader to Section 3.2 and 3.3 of that paper.
\end{remark}

\begin{algorithm}
\caption{Restoring images algorithm}\label{alg}
\begin{algorithmic}
\Require Corrupted image $x_T$
%\Input Corrupted image $x_t$
\For{$t=T, T-1, \dots, 1$}
\State $\tilde x_0= R(x_t,t)$
\State $x_{t-1}=D(\tilde x_0, t-1)$
\EndFor\\
\Return Restored (new) image $\tilde x_0$
\end{algorithmic}
\end{algorithm}

In \cite{cold} the authors showed that this approach works, generating new images, for many different types of deterministic (and random) degradation process such as:
\begin{itemize}
    \item adding deterministic Gaussian noise, which is equivalent to the DPM but the Gaussian noise is fixed, i.e. in \eqref{process} $z_t=z$ established just once and then used at any time step;
    \item blurring images;
    \item animorphis, where human images are transformed in animals pictures;
    \item ...
\end{itemize}
It is interesting, not only the fact that it shows that some similar results to the one in DPM and DS can be obtained with deterministic processes, but also the importance of the limit distribution of the degradated images $x_T$. Indeed, as it is clear from \cite{cold}, one thing that it is in common in all the three models that we presented here is that to generate a new image it is necessary to start from a sample of the limit distribution obtained with the forward processes. In the DPM and DS this limit distribution is simply a $\mathcal{N}(0,I)$ but it is only because the forward processes are Gaussian diffusions. For example in \cite{cold} the limit distribution of animorphis transformations are completely different and are obtained empirically. Then from this empirical distribution is sampled some element and it is restored with the operator $R_{\theta}$ (which is strongly dependent on $D$) built as we described above.

\begin{remark}
    This importance of the limit distribution is addressed really well and precisely in \cite{nongaussian} where the authors proposed different kind of (random) noising process all of them leading to a different limit distribution. The inference algorithm starting point is then always sample an element from one of the limit distribution (obtained empirically or analytically) and ``restore'' it.
\end{remark}

\section{Assessing models' novelty and realism}\label{last}
In the context of DPM it is quite an interesting research question to define ways to assess what levels of realism and novelty  they are able to achieve. We observe that this aspect is normally underestimated and we try here to give a very first hint at how this problem could be approached.

In order to make these sorts of considerations more rigorous, but avoiding at the same time to enter into too philosophical questions - we should obviously start from defining what do we mean by ``existence of an image''. For the sake of simplicity, we can say that an image exists if there exist(ed) an observation event (to be intended in the most general sense, including dreams, imaginations, hallucinations... etc) performed by some observer for that image {\bf{and}} the observer remembers about the event to an extent that it is possible for him/her to recognise the same image. Notice that yet in this simple conceptualisation, we can identify several critical elements which should be further explored and formalised as:
\begin{itemize}
\item{The properties of the set of realistic images $I$, the set of observers $O$ and the set of observation events $E$}
\item{The definition and properties of the image recognition/matching  function $M_o: I\times I\rightarrow [0,1]$ used by an observer $o \in O$ to match two images\footnote{This function is particularly critical since it heavily depends on the specific observer and that includes many features like: the time in which the images are observed, the quantity of times he observed them, etc.}, where, given $i,j \in I$, $M_o(i,j)$ is a  measure of similarity of $i$ and $j$ for the observer $o$ (so that for $i=j$, $M_o(i,j)=1$).  }
\end{itemize}
Therefore, we can say with a certain level of rigorousness that a certain image $x \in I$ is ``new'' if the following condition holds:
\begin{equation}\label{cond}
\nu_O(x):=\sum_{o \in O}\sum_{i\in I_o}M_o(x,i) = 0
\end{equation}
where $I_o \subseteq I$ is the set of images for which observer $o$ recalls an observation event. With \eqref{cond} in mind, we define the set of new realistic images as  $J_O:=\{x \in I: \nu_O(x)=0\} \subset I$. Hence, given a randomly chose image $x \in I$, the probability that it is new, in the sense defined above, is equal to $N_{I,O}=|J_O|/|I|$. This quantity gives account of the intrinsic novelty of $I$ for the community of observers $O$.

A model $M$ is able to produce a certain set of images $I_M=\hat I_M \cup \tilde I_M$ where $\tilde I_M := I \cap I_M$ and $\hat I_M:= I^C\cap I_M$. Indeed, notice that $I_M \nsubseteq I$ in general: some image generated by $M$ could not be ``realistic''. So, the problem of assessing the probability that a generative model $M$ has to generate new images for observers of $O$ boils down to the estimation of how big is the set $I_N\subseteq \tilde I_M \subseteq I$ of new images that can be generated by the model in theory and to evaluate $N_{M,O}=|I_N|/|J_O|$. This quantity gives account of the absolute novelty that images generated by the model show to observers in $O$. If we define $C_M=|\tilde I_M|/|I|$ the completeness of the model, then we have the following fundamental relation:
\begin{equation}\label{novelty}
N_{M,O}:=C_M\frac{N_M}{N_{I,O}}
\end{equation}

where $N_M=|I_N|/|\tilde I_M|$ is the model's relative novelty rate. %\textcolor{red}{questa parte che giustifica \eqref{novelty} è da vedere bene, perché non tiene conto della dimensione di $\hat I_M$ e quindi dell'imprecisione del modello, che potrebbe essere qualcosa tipo $\frac{|\hat I_M|}{|I_M|} \in [0,1]$, per esempio.}

It is becoming clearer how the above definitions make sense only if we are really able to characterise the set $I$. In fact, in practice, when using DPM implementations like \cite{stable} we cannot be sure whether the set $I_M$ is actually a subset of $I$ until we do not characterise $I$. A way to do this is to provide an intensive (synthetic) characterisation $c$, like ``the images of all horses'', and considering then $I_c \subset I$, where $I_c$ is the set of realistic images corresponding to the characterization $c$, instead of the whole $I$. To train $M$, we will then have to provide a sufficient amount of horses pictures and then challenge the model to generate new ones. However, many generated images would probably not resemble horses at all. Therefore to measure the goodness of the model we should establish when an image $x \in I_M$ is in $I_c$. We say that it holds if, at least an observer $o \in O$ would classify it as ``realistic'' and belonging to the intensive concept $c$ that generated $I_c$. Then, defining the boolean classification function $C_{c,o}: I_M\rightarrow \{0,1\}$ as $$C_{c,o}(x):=\begin{cases}
1 & \text{if }  o \text{ recognizes that } x \text{ is realistic and corresponding to } c \,; \\
0 & \text{otherwise}
\end{cases},$$ an image $x \in I_c$ only if
\begin{equation}\label{realism}
\rho(x):=\frac{1}{|O|}\sum_{o\in O} C_{I,o}(x) > 0.
\end{equation}
%%i.e., if there is  at least an observer that recognise $x$ as an example of the intensive definition of $I$.
We can naturally define the overall model realism by: 
\begin{equation}\label{modelrealism}
    R_M:=\frac{1}{|I_M|}\sum_{x \in I_M} \rho(x).
\end{equation}

Working with too generic/generalist concepts like ``all images'', as systems like \cite{stable} do, introduce further complexity to the problem. An approximating approach at this would be started by assuming that for these generalist systems $I$ is composed by the union of a very high number of subsets, each defined by some intensive concept derived by induction from a huge observation process of available data, i.e. $I=\bigcup_{c \in C} I_c$, where $C$ is the set of these classes. Correspondingly, all quantities defined in \eqref{novelty}, \eqref{realism} and \eqref{modelrealism} should be appropriately extended to cover this case (e.g. by weighting the contributions of all classes).
\subsection{Relation to reinforcement learning}
What argued above can be seen as correlated with reinforcement learning whereas we consider realism, novelty or whatsoever other measurable property of generative models as rewards that a system may get from the community of observers. Works have been already started into this direction, e.g. \cite{rlsd}, where authors propose a general framework that automatically adapts original user input to model-preferred prompts through  reinforcement learning based on relevance and aesthetically pleasentness feedback.

\section{Conclusions}
We wrote these notes trying to answer our self the following questions:
\begin{itemize}
\item How is it possible that these models generate ``real''/``realistic'' images?
\item What does it mean for a ``real''/``realistic'' image to be ``new'' and how to measure these concepts in practice?
\end{itemize}

Regarding the first question we hope that this work is a good answer, helping the reader understand better why these models work and on which kind of theory they are based. In particular, we think that these notes present a complete analysis of the mathematical background of the DPM and contextualise the improvements brought by popular tools like Stable Diffusion. As such, they could help (as other similar work, see \cite{google} for example) understand the way these systems work and the fundamental properties on which they rely: the unique limit distribution (not its form, only its existence and uniqueness) for the forward process and the functional form of the backward process. This is even more clear from the analysis of Cold Diffusion in Section \ref{cold_sec}). We therefore hope to have explained and made more understandable to the interested reader the state of the art models for image generation. 

Regarding the second question we do not have yet a complete answer and we think that this can be seen as the main question that generative models are supposed to tackle. We made some interesting considerations about this topic in Section \ref{last} that we resume below.
We can in fact interpret the space of the $x_0$ such that $p_{\theta}(x_0)>0$ as a subspace of the set of all realistic images $I$ which is - by construction - far bigger than the set on which these models are trained. It is intuitively clear that this characterization of the models' output includes more images than all the ones observed in a lifetime by any existing observer too. Under this respect, images generated by DPM can be considered really ``generated'' by them and really ``new''. With this paper we believe to have given a small hint for a step in the direction of characterizing the space of  ``real''/``new'' images, something which is still completely not understood. Our starting point is that we - as observers - are not able to say what defines an image as ``real'' even if we are immediately able to understand if it is such or not by comparing the image through the criteria we formed in our previous experience of observers. This is also the basis by which we are able to asses ``novelty'', i.e., through comparing whether the presented images are resembling anything already experienced in the past. Maybe understanding and developing these model further will help in the future to understand more in detail this quite complex domain and fully characterize it too. We finally like to leave the reader with what may seem a paradox: according to the above interpretation, once an observer has seen an image generated by a DPM, this image instantly loses the status of ``new'' (recalling the criteria defined in \eqref{cond}). Thus, if we make a big enough number of observers watch the results of a big enough run of image generations could we eventually run out of ``new'' images, downscaling what now seems an almost magical property of these systems to generate novelty to a bare capability of representing a distribution of ``existing'' data?

\nocite{*}
\bibliographystyle{plain}
\bibliography{DiffusionModels}

\end{document}